# Spot Evasion Attacks: Adversarial Examples for License Plate Recognition Systems with Convolutional Neural Networks

QIAN Ya-guan[1], MA Dan-feng[1], WANG Bin[2], PAN Jun[1]∗, WANG Jia-Min[1], CHEN Jian-Hai[3], ZHOU Wu-Jie[4], LEI Jing-Sheng[4]

[1](*School of Big-Data Science, Zhejiang University of Science and Technology, Hangzhou 310023, China*)

[2](*Network and Information Security Laboratory of Hangzhou Hikvision Digital Technology Co., Ltd. Hang Zhou 310052, China*)

[3] (*College of Computer Science and Technology, Zhejiang University, Hangzhou 310027, China*)

[4](*School of Information and Electronic Engineering, Zhejiang University of Science and Technology, Hangzhou 310023, China*)

**Abstract**: Recent studies have shown convolutional neural networks (CNNs) for image recognition are vulnerable to evasion attacks with carefully manipulated adversarial examples. Previous work primarily focused on how to generate adversarial examples closed to source images, by introducing pixel-level perturbations into the whole or specific part of images. In this paper, we propose an evasion attack on CNN classifiers in the context of License Plate Recognition (LPR), which adds predetermined perturbations to specific regions of license plate images, simulating some sort of naturally formed spots (such as sludge, etc.). Therefore, the problem is modeled as an optimization process searching for optimal perturbation positions, which is different from previous work that consider pixel values as decision variables. Notice that this is a complex nonlinear optimization problem, and we use a genetic-algorithm based approach to obtain optimal perturbation positions. In experiments, we use the proposed algorithm to generate various adversarial examples in the form of rectangle, circle, ellipse and spots cluster. Experimental results show that these adversarial examples are almost ignored by human eyes, but can fool HyperLPR with high attack success rate over 93%. Therefore, we believe that this kind of spot evasion attacks would pose a great threat to current LPR systems, and needs to be investigated further by the security community.

# 1  Introduction

With the rise of deep learning technology, CNNs have been successfully applied to a variety of image recognition fields, such as vehicle license plate recognition [1, 2, 3], face recognition [4], object detection [5, 6], and so on. In fact, CNNs have been developed to achieve a remarkable performance comparable to that of human eyes, and moreover, in some cases it even outperforms that of human beings [4]. However, recent studies have shown deep neural networks are vulnerable to *adversarial examples* that are carefully modified [7-14]. In general, an adversarial example is an image added by some tiny perturbations in the digital world, which is imperceptible to human eyes, or designed as a resemble graffiti, a relatively common form of vandalism in the physical world [14]. This kind of adversarial examples is malicious, and may lead CNNs to output a fatal misjudgment. Just considering an autonomous driving system, if a road sign are incorrectly recognized, some serious traffic accidents may take place.

---

∗Corresponding author
Email address: panjun78@hotmail.com (PAN Jun)

At present, CNNs based LPRs have been widely used in various scenarios, such as parking lot vehicle registration, road traffic monitoring, and so on. However, a few drivers attempt to evade monitoring by modifying their license plates or even replacing them with fake ones, just to escape from being recognized by automatic monitoring systems when violating traffic rules. Apparently, once identified, they will be severely punished, since the image evidence is so clear and firm. In some cases, however, some drivers would deliberately keep muds or stains on their license plates to deceive LPR systems. This is obviously illegal, yet it would be very difficult for the police to deal with it. Fortunately, such naturally formed spots on license plates cannot deceive LPR systems so easily. However, if the attacker utilize *adversarial examples* with optimization algorithms, it is possible for them to find appropriate positions on the plate for adding spots. We must be aware that such methods might cheat LPR systems with high success rate, and the attacker will thus escape from punishment even their violations are detected by the police, since the evidence is so ambiguous. In this paper, we demonstrate the feasibility of such *spot evasion attacks*, and suggest the necessity and urgency for further research to the AI security community.

In this paper, we limit adversarial examples to images. To the best of our knowledge, there are mainly two different types of adversarial attacks against image classification systems. One approach is to modify all the pixels of images in the digital world, such as the FGSM approach [7], and the other is to add perturbations to some specific regions of physical objects, such as sticker attacks on a traffic sign [14]. Our approach, however, is different from the above. Unlike adding perturbations to the whole image, we introduce local region perturbations as [14] to simulate natural muds or stains on license plates (which often happens on a muddy road). Similarly, unlike sticker attacks with local changed pixels [14], we preset the fixed perturbation value for realistically simulating spots. Therefore, the optimization objective in our problem becomes finding an optimal spot position in an image with fixed perturbation values. As a contrast, in FGSM, it optimizes perturbations for all the pixels in an image, and in a sticker attack, it optimizes perturbations for specific local region pixels.

A spot evasion attack on LPRs belongs to *black-box attacks*, which assumes the attacker cannot fully obtain internal information of the CNN in LPR systems. Conside -ring the transferability of adversarial examples across different CNNs [15], we trained a local substitute CNN as does in [30]. Specifically, we use a publicly available license plate image set to train this substitute CNN. Then we can use the substitute CNN to generate spot adversarial examples of *source characters* in the digital world. After obtaining the spot position of adversarial example, we further add the spot to the source character on license plates according to the obtained spot position, and finally successfully deceive the targeted LPR system in the physical world. The whole procedure is demonstrated in Fig.1.

In this work, the main challenge is to find an optimal position for adding a spot under fixed magnitude of perturbation. We model this as an optimization problem to output the maximum confidence of the *target class* (i.e., the attacker desired class). Therefore, our approach belongs to *targeted attacks*, that is, we attack on a specific source character, and attempt to lead the CNN to classify it incorrectly as other target character (e.g., misclassify a character 'A' as 'F'). Since CNNs are highly nonlinear and non-convex, traditional numerical optimization methods are not effective to find the global optimal position. For this reason, we develop a method based on genetic algorithms to obtain the optimal position. Note, for different target classes, the best position of source character to be modified is also different.

The main innovations and contributions of our work can be summarized as follows:
- We introduce a new approach to generate adversarial examples for attacking the LPR systems. In LPR application scenarios, it is not reasonable to change every pixel value for perturbations. On one hand, the camera cannot capture tiny changes of pixels when transformed to physical objects. On the other hand, the pixels changed with optimization perturbation are not close to real spots. To the best of our knowledge, we are the first to propose spot adversarial examples using optimal perturbation positions, which is not limited to LPR systems as well.
- We simulate a variety of spot forms, including rectangle, circle, oval and spot cluster. The attack effects of different forms of spots are analyzed carefully. We find that each kind of license plate character has a specific attack sensitive region, which provides some interpretability for explaining the working mechanism of CNNs. We also find some characters are more vulnerable to spot attacks, while some characters are suitable for target classes.
- We carry out our experiments on a large number of real Chinese license plate images. The experimental results show that our method can successfully deceive the popular license plate recognition system HyperLPR [16], with a success rate of more than 93%.

The rest of the paper is structured as follows: In section 2, we review different methods to generate adversarial examples including all pixels perturbed and partial pixels perturbed. Section 3 demonstrates the basic principle of LPRs and CNNs. Section 4 describes details of our spot adversarial example generation algorithm. Finally, section 5 describes our experiments and results of spot evasion attacks.

## 2 Related Work

Recent studies have shown that CNNs are vulnerable to adversarial examples [7-14]. To date a variety of methods for generating adversarial examples are proposed, which can divide into two categories. One is to tamper with all the pixels in the whole image in the digital world, and the other is to tamper with the pixels on the local region of images in the physical world. Our approach proposed in this paper belongs to the latter. However, different from previous local region attack methods, we do not optimize pixel perturbations, but try to find an optimal attack positon of images to add a simulated spot with specific color.

## 2.1 Adversarial Examples with All Pixels Perturbed

Szegedy et al. [17] first found that DNNs are vulnerable to adversarial examples, they proposed a box-constrained optimal perturbation method called L-BFGS. Since L-BFGS used an expensive linear search method to find the optimal value, which was time-con -suming and impractical. Goodfellow et al. [7] proposed FGSM (Fast Gradient Sign Method) to generate adversarial examples. They only performed one-step gradient update along the direction of the sign of gradient at each pixel, so the computation cost is extremely lower. However, the generated adversarial example may not perform the best. According to [31], one-step attack like FGSM is easy to transfer but also easy to defend. On the basis of FGSM, many other improved methods are proposed, such as I-FGSM proposed by Kurakin et al. [8], in which momentum is applied to FGSM to generate adversarial examples more iteratively. Tramèr Fet et al. [18] found that FGSM with adversarial training is more robust to white-box attacks than black-box attacks due to gradient masking. They proposed a RAND-FGSM, which added random noise when updating the adversarial examples to defeat adversarial training. The purposes of these methods are to overcome the problem of falling into local optimization in the optimiza -tion process.

Since then, the researchers have proposed a variety of other improved algorithms. Papernot et al. [20] proposed JSMA (Jacobian-based Saliency Map Attack) method for targeted attacks, they perturbed a small number of features by a constant offset in each iteration step that maximizes the saliency map. Compare to other methods, JSMA perturbs fewer pixels. Moosavi-Dezfooli et al. [19] proposed Deepfool to find the closest distance from the original input to the decision boundary of adversarial examples. To overcome the non-linearity in high dimension, they performed an iterative attack with a linear approximation. DeepFool provided less perturbation compared to FGSM and JSMA did. Compared to JSMA, DeepFool also reduced the intensity of perturbation instead of the number of selected features. Su and Vargas et al. [11] proposed a single pixel attack method, which successfully deceive the DNNs by modifying the value of a single pixel. Carlini and Wanger [10] proposed the C&W method to defeat defensive distillation, as far as we know, which is one of the most powerful attacks.

The above attack methods add slight perturbations to the image pixels, and it is not easy for human eyes to detect them. However, if such adversarial examples are converted directly to the physical space (e.g., physical license plates), it will be difficult for cameras to capture them. Therefore, the attack method proposed in this paper is specifically aimed at license plates in physical world, we do not add perturbations to the global pixels of images, but to specific area with predefined spot shapes and colors, which will be easily captured by cameras.

## 2.2 Adversarial Examples with Partial Pixels Perturbed

Adversarial examples crafted with the all pixels of images are usually feasible in the digital world. Therefore, in order to build an adversarial example in the physical world, it is necessary to first locally modify the target object in images. For example, if you want to deceive an autonomous driving system, you are only able to add perturbations to the foreground traffic sign of the image, but not to the background (e.g., the woods behind the traffic sign), for its physically impossibility. Kurakin et al. [8] shows the smartphone camera can be misclassified by the printed adversarial examples in the physical world. They extended FGSM by running a finer optimization for multiple iterations. To further attack a specific class, they chose the least-likely class of the prediction and tried to maximize the cross-entropy. Sharif et al. [12] implemented adversarial eyeglass frames to achieve attack in the physical world: the perturbations can only be injected into the area of eyeglass frames. Mirjali and Ross [21] proposed a technique that can change the gender in a facial photo, while the biometric characteristics of the adversarial example are still in consistent with the original image.

3D printing attacks and patch attacks are another type of attacks. Athalye et al. [13] proposed a method for constructing 3D printed objects that can fool neural networks across a wide variety of angles and viewpoints. Their 'Expectation Over Transformation' (EOT) framework is able to construct examples that are adversarial over an entire distribution of image/object transformations. In our opinion, results of this work ascertain that adversarial attacks are a real concern for deep learning in the physical world. Tom B. Brown et al. [22] proposed an attack to create universal, robust, targeted adversarial image patches in the real world. The patches are universal because they can be used to attack any scene, robust because they work under a wide variety of transformations, and targeted because they can cause a classifier to output any target class. These adversarial patches can be printed, added to any scene, photographed, and presented to image classifiers; even when the patches are small, they cause the classifiers to ignore the other items in the scene and report a chosen target class. Evtimov et al. [14] proposed an attack algorithm, Robust Physical Perturbations (RP2), to modify a stop sign as a speed limit sign. They changed the physical road signs by two kinds of attacks: (1) overlaying an adversarial road sign over a physical sign; (2) sticking perturbations on an existing sign. Our method is different from theirs in that they use the optimization method to calculate the pixel perturbation after selecting the position of the perturbation, while in our approach, the perturbation of license plates is fixed, in other words, the color of the spot is predefined, and the goal is to get optimal perturbation positions. We believe this is more in line with practical application scenarios. First, when the perturbation obtained in the digital world (even the most accurate one) is converted to the physical space, a great distortion will be introduced, due to the influence of physical factors (such as lights, viewpoints etc.). Second, the perturbation color obtained by the optimization algorithm is not consistent with the actual spot color.

# 3  Principles of LPRs

In general, the working principle of LPRs can be divided into the following three steps: (1) locating the position of the license plate in the image; (2) subdividing the characters in the license plate; (3) classifying and recognizing each character. Common methods for license plate character recognition include CNNs [1, 2, 3], SVMs [23], character template [24] and HDRBMs [25]. Among them, CNNs have the ability to extract image features automatically, and generally perform better than other methods in terms of image classification. At present, CNNs are the most widely used models in commercial LPR systems, that is why our proposed method aims to attack Chinese license plate character classifiers based on CNNs.

CNNs are multi-layer feedforward neural network structures, which are mainly composed of convolution layers, pooling layers and full connection layers. The convolution layer contains a set of convolution kernels (also known as filters), which works on a given input image and output a feature map [26]. Convolution kernels are essentially a group of connection weights obtained through backward propagation algorithm, whose function is to extract image features. Pooling layers are typically located behind convolution layers, which compress the information from the convolution layer with typical pooling operators, including maximum pooling and L2 pooling. Full connection layers are usually a typical fully connected neural network structure at the end of CNNs, which output the class predictions of images.

The LPR system in this paper uses CNNs to recognize characters 0~9, A~Z, and the Chinese character representing the province in the license plate (e.g., 浙 (Zhejiang province), 沪 (Shanghai province), 川 (Sichuan province), etc.). In general, multiple separate CNNs are trained respectively to recognize the Chinese characters, the numbers, or the letter characters. Let $f(\cdot)$ denote the CNN classifier for license plate characters, $X$ denote an input of character image, then the predictive output of the CNN is a probability vector $Z = f(X)$, each dimension of which corresponds to the predictive confidence of each possible class, here the prediction label $y$ is an index of class with maximum confidence:

$$y = \arg\max_{i \in 0,...,N-1} Z(i) \tag{1}$$

where $Z(i)$ refers to the $i-th$ class predictive confidence.

## 4 Methodology

Since LPR systems are usually built inside of devices, the attacker usually has no access to internal parameters of systems. Therefore, we assume black-box attacks for our proposed spot evasion attacks. Based on this assumption, a novel method is proposed for generating spot adversarial examples. As shown in Fig.1, we first obtain the license plate image through cameras, and cut out characters from the image. Our purpose is to deceive LPR systems by adding spots to the character. For obtaining the maximum attack success rate, we select the rest characters as candidate target classes, and try to manipulate the source character (e.g., 'D'), and output its all of the target class prediction confidences (e.g., target class A is 0.723, B is 0.941 etc.). As a result, the target class with the maximum confidence is considered as the final attack target class for this source character. For reducing the number of spots and achieving better camouflage effect, we merely select one source character to add spots. Therefore, we further determine the source character with the maximum misclassified confidence over all candidate source characters. In this example, we finally select 'D' as a source character since its corresponding target class 'B' has the maximum misclassified confidence. After obtain the adversarial example in digital world, we can further add spots to physical license plate according to the indicated position. Note, in the whole procedure, the key component is to find the optimal spot position for guarantee of the maximum misclassified confidence, which we will describe in Sec 4.3 in detail.

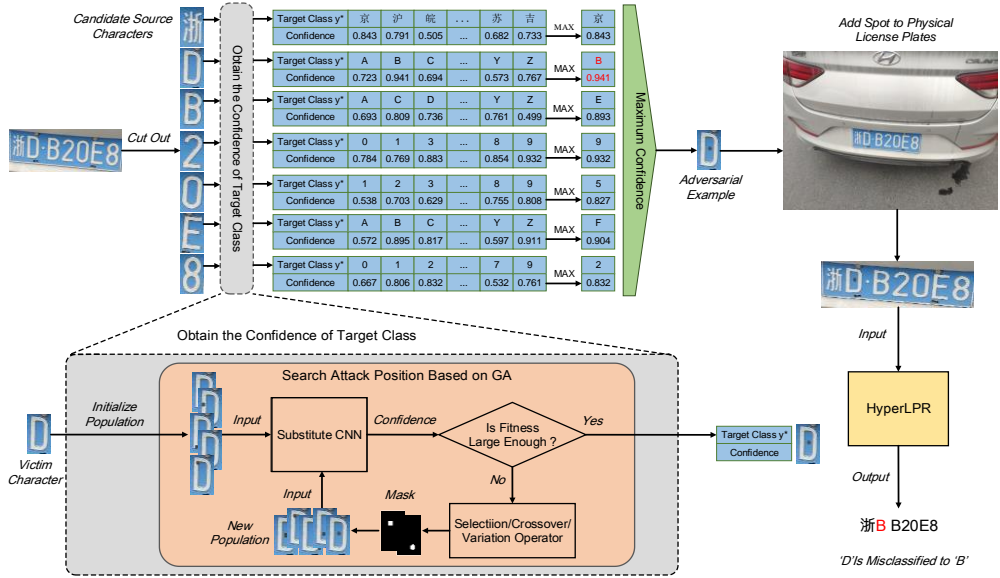

Fig 1: The procedure of spot attacks on license plates

## 4.1 Threat Model

Yuan et al. [27] presented a deep learning threat model in two dimensions. The first dimension is the adversarial goal including *targeted attacks* and *untargeted attacks* according to the adversarial specificity. The second dimension is the attacker ability defined by the amount of information that the attacker can obtain from target CNNs, which are divided into two categories (i.e., *white-box* attacks and *black-box* attacks).

**Adversarial goal**: Let $f(\cdot)$ denote the CNN classifier, $X$ denote the input example with a ground-truth label $y$, and $X'$ denote the generated adversarial example from $X$. For untargeted attacks, the goal is to lead the CNN to classify $X'$ into any other classes (i.e., $y' \neq y$, where $y' = \arg\max Z'(i)$, $i \in 0,...,N-1$), $Z' = f(X')$. While for targeted attacks, the goal is to lead the CNN to classify $X'$ to the attacker desired class $y^*$ (i.e., $y' = y^*$ where $y^* \neq y$).

Generally speaking, targeted attacks are more difficult than untargeted attacks [27]. In this paper, we aimed at targeted attacks. Specifically, we add some perturbations to the license plate character image $X$ with ground-truth label $y$, simulating the real spots (such as muds, etc.), and attempt to generate the adversarial example $X'$. The objective is to lead the license plate character classifier CNN to predict the label as $y^*$.

**Adversarial capabilities** are defined by the amount of information the attacker has about the target classifier. The so-called white-box attack means that the attacker has almost all the information of the neural network, including training data, activation functions, network topologies and so on. The black-box attack, however, assumes the attacker has no way to obtain the internal information of the trained neural network model, except the output of the model including label and confidence. Under white-box attack assumption, although it is more likely for the attack to be successful, in general, it is not feasible in reality. However, for black-box attacks, there is no need for the attacker to know the internal information of the model, so it is more in line with real-world attack scenarios, such as LPRs. Therefore, we assume the attack on the LPR is a black-box attack.

## 4.2 Problem Formalization

As shown in Fig.1, we denote the input character image as $X$, and the predictive confidence of the label $y$ of $X$ as $f_y(X)$. Previous studies treated perturbations as a matrix variables $\sigma$, the same size of the input, and denoted the adversarial examples as $X+\sigma$, just as Fig.2(a) shows. In that case, the goal is to lead the classifier to predict the perturbed image as $y^*$, which can be modeled as the following optimization problem [7-11]:

$$\arg\max_{\sigma} f_{y^*}(X+\sigma) \quad \text{s.t.} \quad \|\sigma\|_p < C \tag{2}$$

The objective of Eq.(2) is to find an optimal perturbation $\sigma$ under some constraints, in general $p-norm$. Different from that, we propose an adversarial example attack that simulates the local spot of license plate character. We preset the perturbation value, that is, the color of the spot is predetermined, and then search for a best perturbation position in the original image, and replace it with the spot, so as to generate an adversarial example that can successfully deceive the LPR, as shown in Fig.2(b).

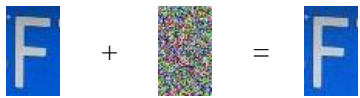

(a)Adversarial examples of perturbation attacks

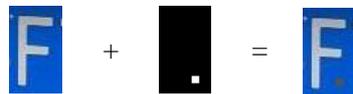

(b)Adversarial examples of spot evasion attacks

Fig. 2: Illustration of the difference between two types of adversarial examples

Fig.2 intuitively shows the difference between our proposed spot adversarial examples and previous adversarial examples with pixels perturbed. In our method, a spot position is a variable to be optimized, with a preset perturbation value (spot color), while in previous adversarial examples, adversarial perturbation of the whole or partial image is a variable to be optimized. Therefore, in our problem, the generation of adversarial examples can be modeled as an optimization problem which aims to find an optimal spot position. We introduce the concept of mask to represent the shape and size of spots. Formally, a mask is a matrix $M$, in which the region that has no spot is represented as 0, and that has a spot is represented as 1. In this way, by setting a position to 1 in the mask matrix, we can present a spot position and its shape. For a rectangular spot, $M(a,b,r)$ represents it with width $r$, starting from the point $(a,b)$ and ending at the point $(a+r,b+r)$ limited in the original image. Suppose the scalar $\delta$ is the amplitude of the perturbation, we can obtain the spot perturbation from the mask $M(a,b,r)\cdot\delta$, which is corresponding to the parameter $\sigma$ in the Eq.(2). Therefore, the above process can be further modeled as the following constrained optimization problem:

$$\arg\max_{a,b} f_{y^*}(X \odot (J - M(a,b,r)) + M(a,b,r)\cdot\delta)$$
$$s.t. \quad a \in [0, X_a - r], b \in [0, X_b - r] \quad (3)$$

where $X_a$, $X_b$ represent the length and width of the original image, $J$ denotes the all-1 matrix with the size of $X_a \times X_b$, $X \odot (J - M(a,b,r))$ denotes the Hadamard product of the original image and the mask matrix, which will set the perturbed area to all zero, and the constraint will restrict the spot to the original image area.

### 4.3    Optimization with Genetic Algorithm

We model a generation procedure of spot adversarial examples as a constrained optimization problem, however, under the assumption of black-box attacks, parameters of CNNs are not accessible, and we cannot obtain gradients of position to objective functions. So we use a genetic algorithm (GA) to solve Eq.(3). GA is a random search method based on biological evolution, which has the following advantages in solving our optimization problem.

- It is easier to find global optimal points. Previous attack methods, such as FGSM, C&W and RP$_2$, all attempt to solve the problem by gradient descent approach, which make it easy to fall into local optimal points when dealing with multi-valley objective functions of CNNs. On the other hand, GA adopts global random search, which avoids falling into local optimal points, and converges to global (or approximately) optimal solution finally.
- There is no need to obtain gradient information of CNNs. White-box attack need to obtain gradient information to compute perturbations, which makes it unavailable under a black-box assumption. However, under a black-box assumption, GA can solve Eq.(3) without gradient information.

GA is an imitation of the natural selection mechanism through 'survival of the fittest'. Consider the optimization problem as follows:

$$\max g(x) \quad s.t. \quad x \in \Omega \quad （4）$$

Where $\Omega$ denotes a feasible set, $x$ represents a chromosome of $\Omega$, and $g(x)$ is a fitness function of chromosomes. The basic idea of GA to solve the optimization problem can be described as follows: first, a group of initial chromosomes (also known as initial population) are randomly selected from the feasible set $\Omega$, and then the fitness of these chromosomes $g(x)$ is calculated. Finally, these chromosomes (parents) are crossed and mutated to produce the new chromosomes (offspring) $P^{(1)}$. The above steps are repeated and the new populations $P^{(2)}$, $P^{(3)}$, …, are produced, until the stop condition (generally the iteration number) is satisfied. The purpose of crossover and mutation operation is to create a new population, so that the average fitness of the new population will be greater than that of the previous generation.

We will use GA to solve the optimization problem described in Eq.(3). In Eq.(3), a position of spot $(a,b)$ is a decision variable in the feasible set, which represents a chromosome. Here a chromosome is binary coded to represent a possible spot position. $a \in [0, X_a - r], b \in [0, X_b - r]$ represents the feasible set $\Omega$, which means that the optimization process needs to be limited within the image. We take the prediction confidence $f_{y^*}(\cdot)$ of CNNs as the fitness of GA. Therefore, in this problem, the maximum fitness of GA is exactly the maximum confidence of target class. The detailed implementation procedure is shown in Algorithm 1.

---

**Algorithm 1：Searching attack position**

**Input：** character image $X$，$r$：rectangular spot side length，$J$：all-1 matrix with $X_a \times X_b$  $\delta$：amplitude of perturbation，$y^*$：attack target class，$x_i$：chromosome

**Output：** the optimal perturbation position $(a^*, b^*)$

---

1：Initialize $P_c = 0.8, P_m = 0.01, PS = 10, k = 0, G = 100$, and $T_f = 0.99$, where $P_c$ is the cross probability，$P_m$ is the variation probability，$PS$ is the population size，$G$ is the generations of evolution，$T_f$ is the threshold of fitness

2：Calculate the encode length of the chromosome using $dv = [[0, X_a - r], [0, X_b - r]]$, where $dv$ is the spot location range

3：Randomly generated population $P^{(0)} = \{x_i, i = 1, ..., PS\}$, where the chromosome $x_i$ is randomly coded according to encode length

4：**do**

5：　　$(a_i, b_i) \xleftarrow{decode} x_i,\ i = 1, ..., PS$

6：　　compute $f_{y^*}(X_i'),\ X_i' = X \odot (J - M(a_i, b_i, r)) + M(a_i, b_i, r) \cdot \delta,\ i = 1, ..., PS$

7：　　$P^{(k+1)} = \varnothing$

8：　　**while** ($|P^{(k+1)}| < PS$) **do**

9：　　　　use roulette-wheel selection according to $f_{y^*}(X_i')$ to select $x_m$ and $x_n, m, n \in [1, ..., PS]$

from $P^{(k)}$

10：　　　　　**if** ( $random(0,1) < P_c$ ) **then**

11：　　　　　　$x_m$ and $x_n$ conduct one-point crossover operator with probability $P_c$ to generate new chromosomes $x'_m$ and $x'_n$

12：　　　　　　$x_m \leftarrow x'_m$,　$x_n \leftarrow x'_n$

13：　　　　　**end if**

14：　　　　　**if** ( $random(0,1) < P_m$ ) **then**

15：　　　　　　$x_m$ and $x_n$ conduct variation operator with probability $P_m$ to generate new chromosomes $x'_m$ and $x'_n$

16：　　　　　　$x_m \leftarrow x'_m$,　$x_n \leftarrow x'_n$

17：　　　　　**end if**

18：　　　　　add the new chromosomes $x_m$ and $x_n$ to $P^{(k+1)}$

19：　　　**end while**

20：　　$P^{(k)} \leftarrow P^{(k+1)}$

21：　　$k \leftarrow k+1$

22：**while** ( $f_{y^*}(X'_i) < T_f$ **or** $k < G$ )

23：$(a_i, b_i) \xleftarrow{decode} x_i$, $x_i \in P^{(k)}$, $i = 1, ..., PS$

24：$a^*, b^* = \underset{a_i, b_i}{\arg\max}\, f_{y^*}(X'_i)$, $X'_i = X \odot (J - M(a_i, b_i, r)) + M(a_i, b_i, r) \cdot \delta$, $i = 1, ..., PS$

25：**return** $(a^*, b^*)$

### 4.4　Designing Spot Shapes

In order to make an adversarial spot look more naturally, we designed it in three forms: rectangle, circle, and spot cluster.

**Rectangular spots.** In previous studies, a perturbed region is usually modeled as a box constraint $a \in [0, X_a - r], b \in [0, X_b - r]$ representing a rectangular spot in geometry [14]. The spot can be demonstrated by a rectangle area on mask $M$. In experiments, we set the width $r$ and perturbation $\delta$ to different values, and then compare attack effects of the adversarial example. Specifically, we set $r$ to 3, 5 and 7 pixels respectively, since the size of spot needs to be limited within a reasonable range, that is to say, if the spot is too large, it will be easily detected by human eyes, and will be considered as intentionally manipulating the license plate. Nevertheless, if the spot is too small, it cannot deceive LPR systems effectively. The perturbation range $\delta$ is limited within a range of [-255, 255], and $\delta = \pm 50, \pm 100, \pm 150, \pm 200, \pm 255$. Considering the perturbation has three RGB color channels, just as in the original image, the values for all three RGB color channels are set to the same value in the experiment.

**Circular spots.** Although adversarial examples with rectangular spots reach a high attack success rate, they in license plates do not look so natural and real. To this end, we use circular or ellipse spots instead, which can imitate the "mud spatter on the license plate" scene in real scenario more naturally. With an appropriate size, we find that circular spots will not only avoid being noticed by human eyes, but also deceive LPR systems effectively. A circular spot is represented by the following constraint in Eq.(3): $a \in [r, X_a - r]$, $b \in [r, X_b - r]$, where $a, b$ are the coordinates of circle center. For avoiding duplicating the experimental results from rectangular spots, we only select the radius $r = 3$ and the perturbation value is the same with that of the rectangular spot. For elliptical perturbations, the major radius $r_1 = 3$, and the minor radius $r_2 = 2$, and the constraint is represented as $a \in [r_1, X_a - r_1], b \in [r_2, X_b - r_2]$.

**Spot cluster.** In addition to the above simulation of a single spot, we also simulate a cluster with multiple spots to approximate reality. The specific design of spot clusters is shown in Fig.3. Several spot positions obtained by Algorithm 1 are overlapped in the same character image. The overlapping part of more than three spots is considered as cluster positions, so that a mask of the spot cluster would be obtained. Finally, the adversarial example is obtained using the new mask.

**Table 1**: Spot positions for successful attacks with source-target pairs

| A→F | B→D | C→A | D→B | E→F | F→E |
|---|---|---|---|---|---|
| (7,23) | (23,27) | (47,2) | (27,19) | (12,23) | (52,11) |
| / | (9,7) | (46,24) | (24,14) | (10,25) | (43,24) |
| / | (28,26) | (18,17) | (27,17) | (31,0) | (46,11) |
| / | (4,0) | (18,14) | (24,14) | (19,25) | (43,22) |
| / | (52,27) | (42,1) | (25,14) | (13,22) | (45,10) |
| / | (34,1) | (15,13) | (20,19) | (11,24) | (43,27) |
| / | (35,12) | / | (24,14) | (12,21) | (46,19) |
| / | (10,1) | / | (28,19) | (13,21) | (15,3) |
| / | (17,24) | / | / | (11,18) | (49,17) |
| / | / | / | / | / | (43,22) |

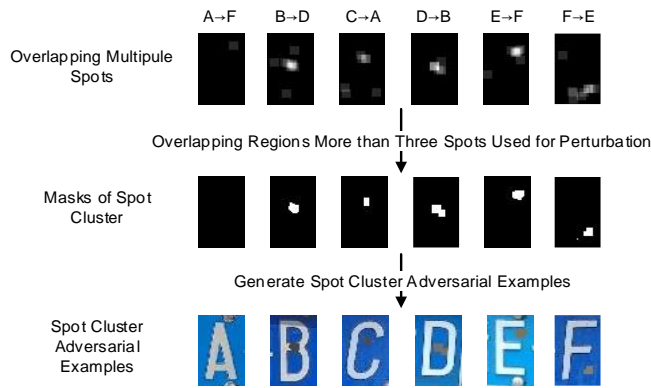

Fig.3 Spot cluster generation: Overlapping Multiple Spots → Overlapping Regions More than Three Spots Used for Perturbation → Masks of Spot Cluster → Generate Spot Cluster Adversarial Examples → Spot Cluster Adversarial Examples

Fig. 3: The generating procedure of adversarial examples with spot clusters

# 5 Experimental Evaluation

To evaluate the performance of our method, we launch attacks against the real license plate recognition system HyperLPR [16]. HyperLPR is nowadays widely used in intelligent parking lot, intelligent transportation and other scenes, and is well known for its high performance. The CNN network structure adopted by HyperLPR is as follows: the input layer is a license plate character image with a size of $35 \times 60 \times 3$, the middle layer is composed of three convolution layers and two pooling layers, using a $5 \times 5$ convolution kernel and a $3 \times 3$ maximum pooling layer. The final layer is a full connection output layer. To evaluate the effect of adversarial examples with different forms of spots, we use two performance measures (i.e., attack success rates and number of targeted attacks).

## 5.1 Datasets

We collect license plate character images through mobile phone camera (accounting for more than 90% of the data) and public available images to train the substitute CNN. After an unified process of image cutting, screening, data enhancement and other operations, we build a training dataset $D$, containing 2300 license plate character images, each with the size of $35 \times 60$. Among them, there are about 380 images of characters 'A' to 'F'. The accuracy of the CNN trained by this data set can reach 98.38%.

For simplicity, in experiments we extracted sixty character images from $D$ that are correctly classified by the CNN as source character set $D_{source}$ to craft adversarial examples, each character has ten images as shown in Fig. 4. These images are taken from mobile phones and have been clipped, screened and enhanced. All the character images in experiments are of good quality to ensure the effect that spots can affect classifiers.

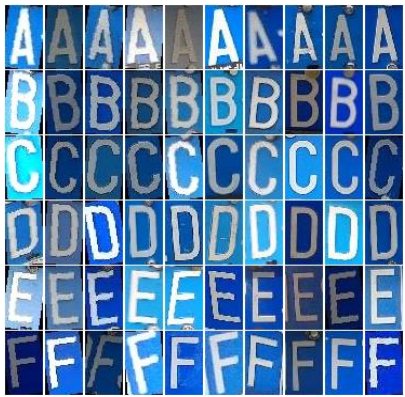

Fig. 4: The source character set $D_{source}$

## 5.2 Evaluation Measures

Since magnitudes of perturbations are strictly limited, not all source images can produce effective adversarial examples. For evaluating the effectiveness of the proposed method, we use two evaluation indicators: attack success rates and the number of successful targeted attacks. We select source characters that have been correctly classified as adversarial examples, so as to make it clear that classification errors are caused by spot perturbations.

**Attack success rates (ASR):** the ratio of the number of images that succeed in attack to the number of source images:

$$ASR = \frac{\sum_{X \in D_{source}} \mathbb{I}(f(X+\delta) \neq y \wedge f(X) = y)}{\sum_{X \in D_{source}} \mathbb{I}(f(X) = y)} \tag{5}$$

where $D_{source}$ is a set of source character images for testing the attack success rate, and $\mathbb{I}(\cdot)$ is an indicator function. $X+\delta$ is an adversarial example with spots, and only when it is misclassified that the attack can be considered successful.

**Number of successful targeted attacks (NSTA):** the number of images that are successfully predicted as other target classes.

$$NSTA = \sum_{X \in \{f(X)=y\}} \mathbb{I}(f(X+\delta) = y*) \tag{6}$$

Where $y$ is the ground truth label of $X$, and $y*$ is the attack target. For example, $NSTA_{(A,E)} = 8$ means that eight of ten character images 'A' are successfully classified as the target character class 'E'. We select six classes of characters from 'A' to 'F', and assume that each class of characters can be attacked and recognized as other five classes, then draw thermal diagram of the number of success attacks against these different target characters.

## 5.3 Experimental Results

We generate three different forms of spots from $D_{source}$ to evaluate the effectiveness of our crafted adversarial examples. In the first experiment, the rectangular spots through our method can always mislead the CNN to predict one class of characters to target class with 100% ASR. In the second experiment, the circular and elliptical spots are crafted to simulate real stains, which have natural camouflage and guarantee high ASR at the same time. In the final experiment, from the view of model interpretability, we try to explore the reason why one class source character attacked into another target class character will have relatively centralized best attack positions.

**Rectangular spots.** The adversarial examples with rectangular spots generated by our approach have the highest ASR up to 93.33%. In Fig.5, we observed that when the perturbation was constant, larger spot size would lead to higher ASR. For example, intuitively, when we set the perturbation $\delta=200$, ASR reaches 90% with a spot size of $7 \times 7$, but drops to 61% when the spot size is reduced to $3 \times 3$. At the same time, we also observed that ASR became higher when the perturbation increased in a positive direction than in a negative direction. This is in consistent with the experimental results of Papernot et al. [20], which shows that reducing pixel intensity is less likely to achieve some expected performance, and the authors explained that information entropy was reduced when pixel numbers dropped, making it hard for CNNs to extract necessary information to classify examples. In addition, when the perturbation value is greater than 150, for adversarial examples with different sizes of spots, ASR begins to slow down (e.g., the success rate of $3 \times 3$ spot cannot exceed 70%). Therefore, with a predetermined perturbation value, we recommend the size $5 \times 5$ for spot adversarial examples.

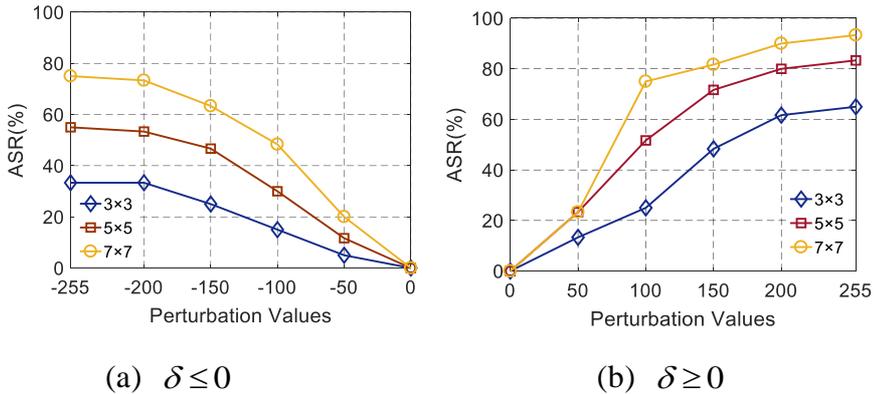

(a) $\delta \leq 0$          (b) $\delta \geq 0$

Fig. 5 ASR of adversarial examples with different spot sizes and perturbation values

Fig.10, 11 and 12 demonstrate the number of successful targeted attacks defined in formula (6), with different spot sizes and perturbation values. The results are represented by thermal maps, where characters on the left belong to the source class and characters on the bottom belong to the target class. The default diagonal value is 10, which means that there are ten images for each source character class, and it also indicates that there are no targeted attacks occurring at matrix diagonal characters. We observe from thermal maps that success rates are different for the same character class misclassified to other target character classes. For example, in Fig.10(g), with the spot size of $3 \times 3$ and perturbation value $\delta = 200$, NSTA of character images 'C', which is misclassified as 'A', is 2, and misclassified as 'E' is 9. This demonstrates that the degree of attacking difficulty is diverse among different target classes for one specific source character image. Therefore, to guarantee a high ASR, we always choose the target class with least degree of attacking difficulty.

We can also observe from thermal maps that the source character with lager row sum will be more vulnerable to spot attacks, while the target character with smaller column sum will be properly considered as targeted class by other source characters for higher ASR. Therefore, under a fixed spot size, we plot the row mean (or column mean) of NSTA with different perturbation values, as shown in Fig.13, 14, 15 and 16. In Fig.13, it shows that the row mean of the source character 'C' is the largest, while that of the character 'A' is the smallest in those source characters, which indicates that 'C' is the most vulnerable, while 'A' is more robust to spot attacks when $\delta > 0$. In Fig.14, it shows that both 'B' and 'C' are vulnerable to spot attacks, but 'A' is still more robust when $\delta < 0$. Fig.15 and 16 demonstrate the NSTA mean of a specific target class from total source characters (i.e., the column mean of a specific target character in the thermal maps). We find that the column mean of target character 'E' is the largest, indicating that 'E' is most likely to be targeted class for other source characters. When $\delta > 0$, the column mean of 'C' is the smallest, indicating that 'C' is less likely to be considered as an attack target. When $\delta < 0$, the character 'B' has the smallest column mean, indicating that 'B' is less likely to be selected as an attack target. Papernot et al. [20] believe that this may be due to the fact that for some input classes, CNNs is easier to classify them. Therefore, in real attack scenarios, the attacker will prefer adding spots on vulnerable source characters such as 'A', and choose target characters with larger NSTA such as 'C' to ensure a high ASR.

**Circular spots**. We further design spots in the form of circle or ellipse shape for more natural simulation, and find that their ASRs reach to 90% and 86.67% respectively shown in Table 2. In Table 2 and 3, ASR of circular spots is very close to that of rectangle spots, if they share similar pixel densities. This indicates that ASR depends mainly on the size and position of spots, but not be relevant to their shapes. Therefore, in practice, we can first obtain an optimal perturbation position under a box-constraint with rectangle spots, and then modify it into circular or ellipse spots, so as to improve its camouflage. Fig.6 demonstrates adversarial examples with circular and ellipse spots. Circular spots and rectangular spots are basically identical in terms of vulnerabilities, so we do not cover this part here.

**Table 2**: ASR of adversarial examples with circular, elliptic and rectangular spots

| Spot Forms | Spot pixel values | | | | | | | | | |
|---|---|---|---|---|---|---|---|---|---|---|
| | -255 | -200 | -150 | -100 | -50 | 50 | 100 | 150 | 200 | 255 |
| 7×7 | 75% | 73.33% | 63.33% | 48.33% | 20% | 23.33% | 75% | 81.67% | 90% | 93.33% |
| Circle | 66.67% | 68.33% | 60% | 50% | 16.67% | 23.33% | 73.33% | 81.67% | 88.33% | 90% |
| ellipse | 58.33% | 55% | 53.33% | 35% | 13.33% | 21.67% | 60% | 78.33% | 86.67% | 86.67% |
| 5×5 | 55% | 53.33% | 46.67% | 30% | 11.67% | 23.33% | 51.67% | 71.67% | 80% | 83.33% |

**Table 3**: Spot sizes of different shapes and their percent in the source image

| Spot Forms | spot pixels | Percent of spot pixels |
|---|---|---|
| Rectangular (7×7) | 49 | 2.33% |
| Rectangular (5×5) | 25 | 1.19% |
| Rectangular (3×3) | 9 | 0.42% |
| Circle | 42 | 2.00% |
| ellipse | 31 | 1.48% |

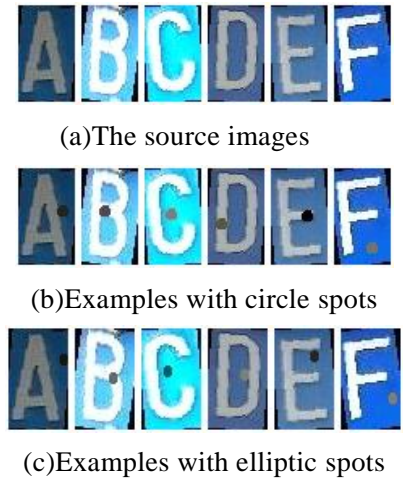

(a)The source images

(b)Examples with circle spots

(c)Examples with elliptic spots

Fig 6: Illustration of adversarial examples with circular and elliptic spots

**Spot Cluster.** Fig.7 shows the mask of spot clusters produced by a single spot, with which about 220 adversarial examples are generated (22 valid masks, each corresponding to ten adversarial examples), and the whole ASR reaches to 63.18%. For rectangular spots, as mentioned above, although a high ASR will be obtained by accurately search the optimal spot position for a specific character image, it also brings a high computational cost. Unlike single spot attacks, spot cluster attacks utilize the rectangular spot template obtained previously, and directly add certain spots to a specific character image. Although the whole ASR of spot cluster is lower than that of single rectangular spot, it is possible for the attacker in practical to generate spot adversarial examples on a large scale due to its lower computational cost.

NSTAs of spot clusters with source-target pairs are demonstrated in Fig.8. Compared with the NSTA of a single spot in Fig.10-12, a spot cluster is higher than a single spot. In Fig.8 (excluding diagonal lines), there are seven source-target attacks to be fully successful (i.e., 10 NSTAs for each of seven source-target attacks, which is more exceeds that of a single spot with ASR more than 90% in Fig.12(i)). The results show that there are more source-target character pairs can achieve a high ASR with spot clusters (e.g., C→B, C→D or C→E and so on). Therefore, in the practical attack scenario, for reducing the computational cost and guaranteeing a higher success rate in targeted attacks, the attacker will directly use the spot cluster to generate adversarial examples.

In Fig.9, we found that there is a strong semantic correlation between spot positions and attack target classes. For example, the character 'D' with a spot added to its middle is closer to the character 'B' in semantics. Fig.7(a) presents an overlapping form of multi-spots. The whitest part of the overlapping region indicates a high overlapping degree (i.e., the most centered attack position, which implies that there is a strong semantic correlation between spot positions and target characters). From the view of a specific attack target, there are relative fixed optimal spot positions for the algorithm to select for source characters due to semantic correlation.

From the generating procedure of spot clusters, we conclude that each character image has their specific sensitive regions that should be important features for neural networks to distinguish characters. Through the study of spots cluster attacks, to some extent, it provides some explanations for the internal working mechanism of neural network, and reconfirms the conclusion given by literatures [28, 29] that neural network is sensitive to some regions in the image, which contributes directly to classification.

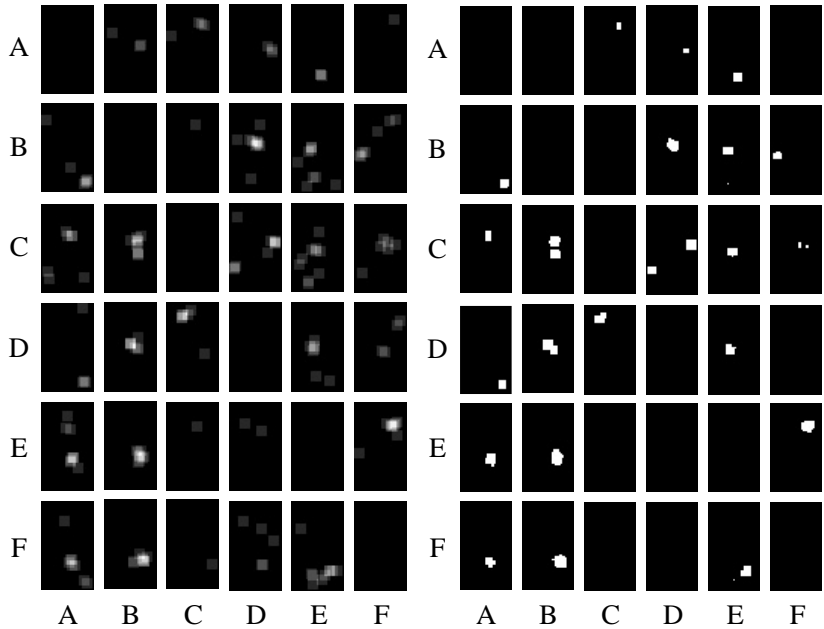

(a) Overlapping of single spots      (b) Mask of spot clusters

Fig. 7: The spot clusters of different target classes where vertical letters representing the source characters, and horizontal letters representing the target classes. (a) overlapping of single spots that succeeds in the attack, with lighter color representing greater overlapping degree; (b) overlapping of more than three spots in (a) is viewed as the spot position of mask.

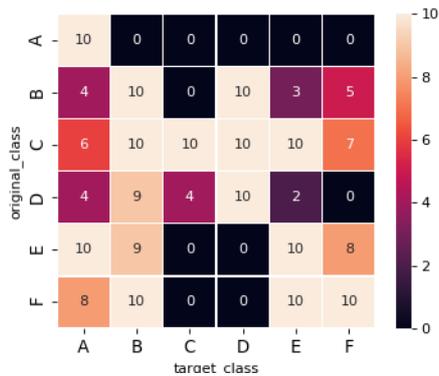

Fig 8: The thermal map of NSTA with spot clusters

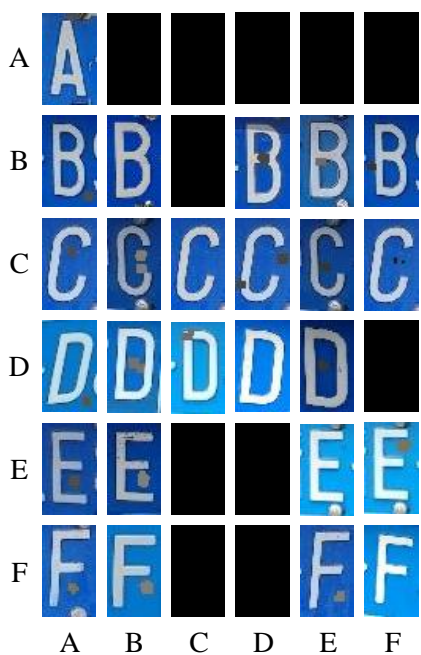

Fig 9: Adversarial examples with spot clusters that succeed in the attack, the black one means no successful adversarial examples generated

# 6  Conclusion

In the previous work, pixel-level perturbations are added to the whole or part of images, and for obtaining tiny perturbation values, various optimization algorithms are used to generate adversarial examples that are close to original images and thus deceive CNNs. However, our work is to generate adversarial examples for physical license plates, which presents obvious perturbations to be captured by cameras. We set fixed perturbation values and size to simulate spot, which can finely simulate natural muds and avoid exciting police attentions. We model this attack as a new optimization problem that attempts to find the best attack position but not pixel perturbation values as previous work. A genetic algorithm is used to solve this hard optimization problem. The experimental results show that the adversarial examples generated by our approach have higher camouflage abilities and attack success rates.

We will extend our work to license plate detectors in the future, which is more complex than classifiers. Meanwhile, we will further take into account the effects of surrounding conditions such as light, shooting angle and visibility degree to generate more adaptive adversarial examples for research community.

**ACKNOWLEDGMENTS**
This work is supported by National Key R&D Program(2018YFB2100400), Zhejiang Provincial Natural Science Foundation of China (No.LY17F020011, No.LY18F020012), Natural Science Foundation of China (No.61672337, 61972357), and Zhejiang Key

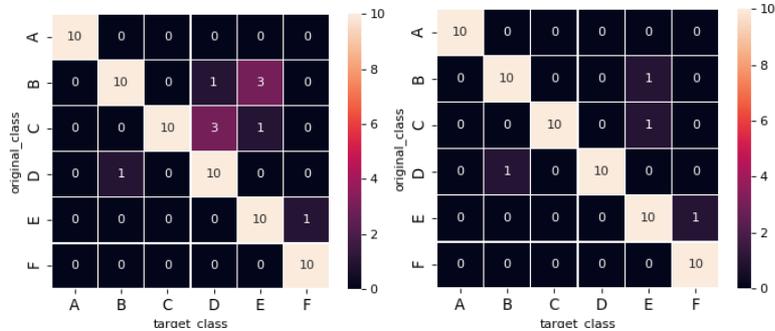

(a) 3×3, $\delta = 50$    (b) 3×3, $\delta = -50$

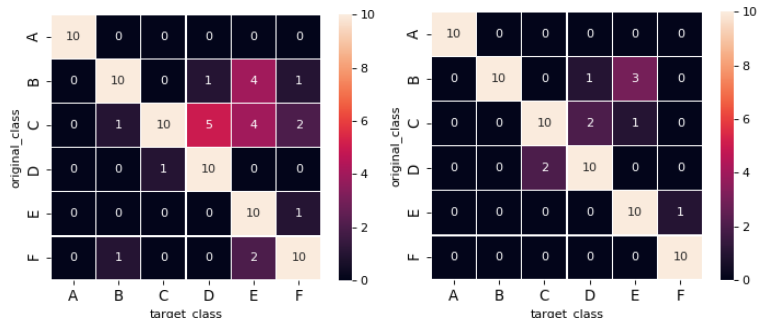

(c) 3×3, $\delta = 100$    (d) 3×3, $\delta = -100$

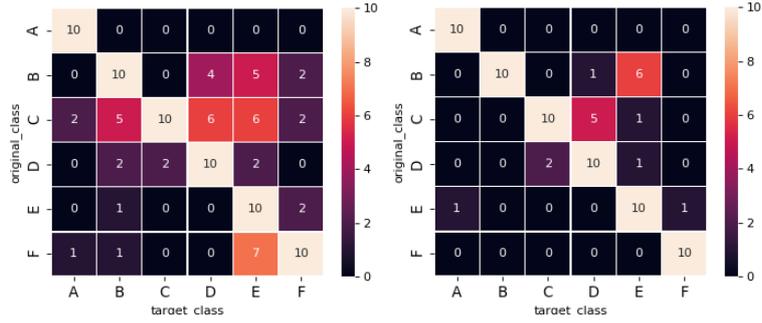

(e) 3×3, $\delta = 150$     (f) 3×3, $\delta = -150$

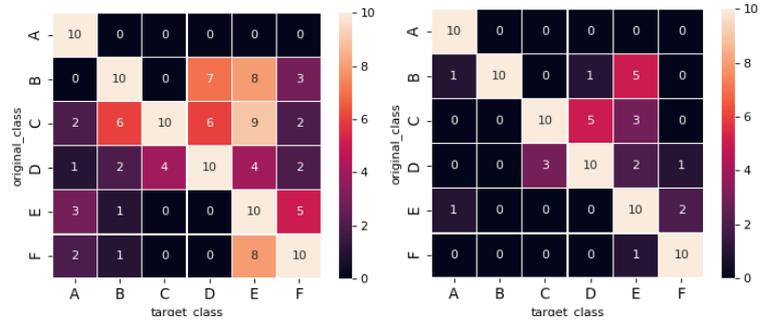

(g) 3×3, $\delta = 200$     (h) 3×3, $\delta = -200$

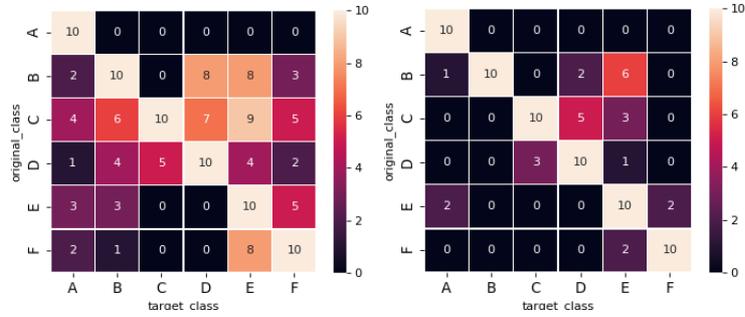

(i) 3×3, $\delta = 255$     (j) 3×3, $\delta = -255$

Fig. 10 Thermal maps for NSTA with 3×3 spots and different perturbation values

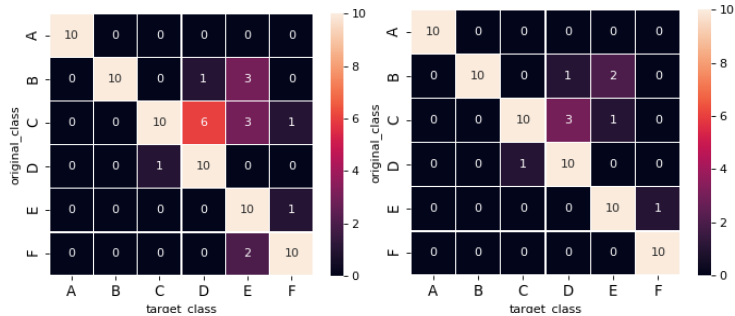

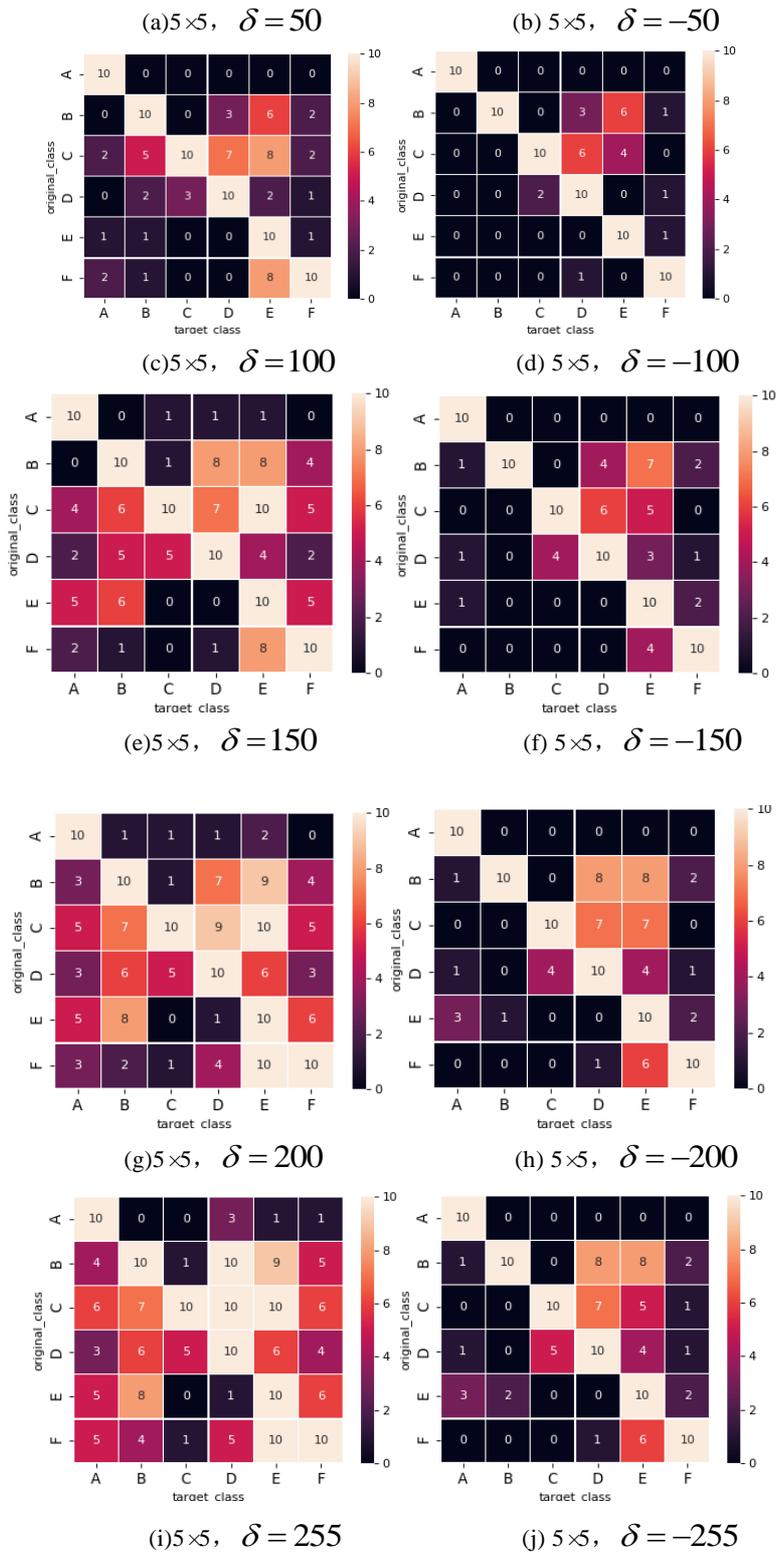

Fig. 11 Thermal maps for NSTA with 5×5 spots and different perturbation values

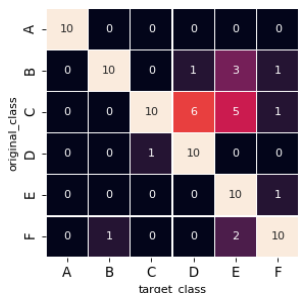
(a) 7×7, $\delta = 50$

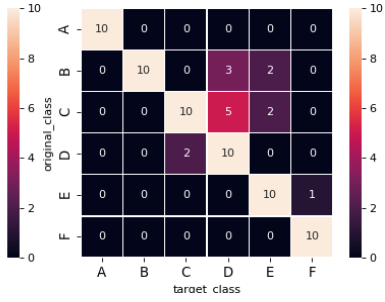
(b) 7×7, $\delta = -50$

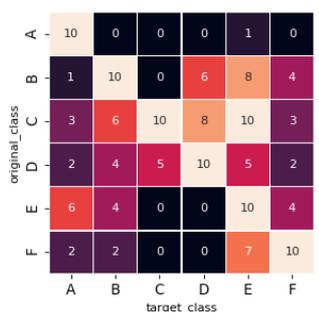
(c) 7×7, $\delta = 100$

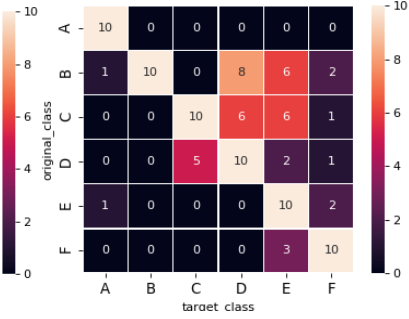
(d) 7×7, $\delta = -100$

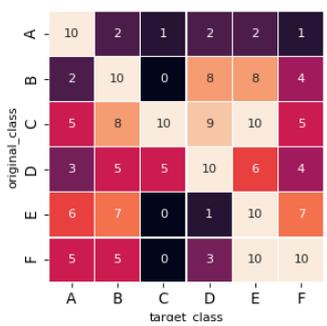
(e) 7×7, $\delta = 150$

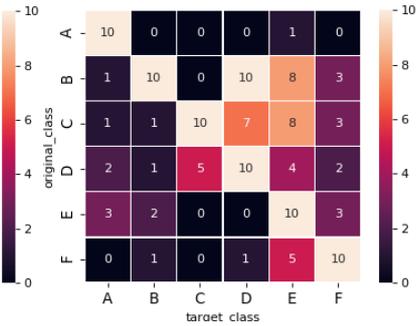
(f) 7×7, $\delta = -150$

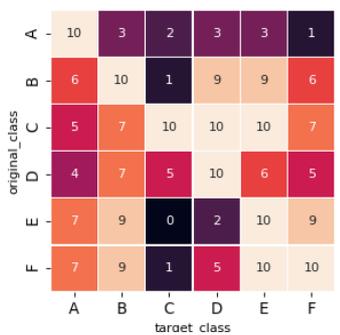
(g) 7×7, $\delta = 200$

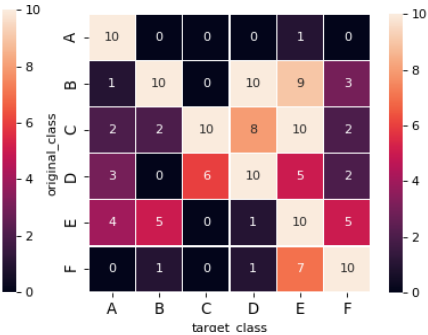
(h) 7×7, $\delta = -200$

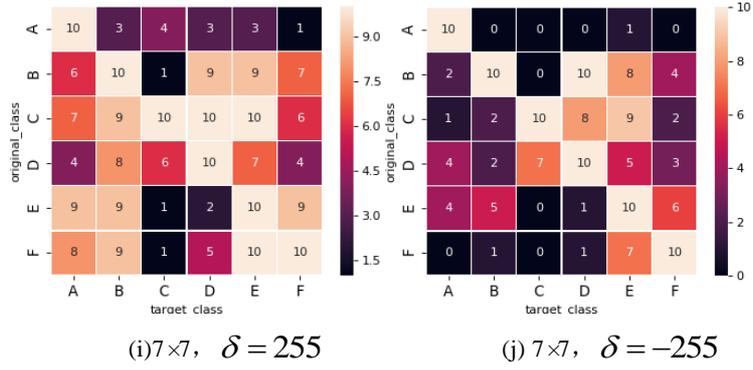

(i) 7×7, $\delta=255$      (j) 7×7, $\delta=-255$

Fig. 12 Thermal maps for NSTA with 7×7 spots and different perturbation values

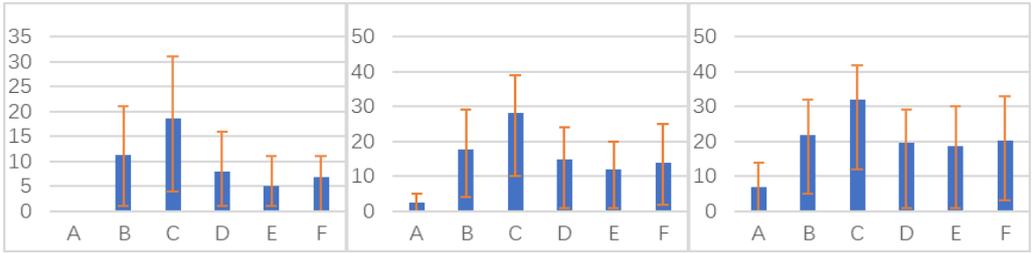

(a) 3×3, $\delta>0$, row sum    (b) 5×5, $\delta>0$, row sum    (c) 7×7, $\delta>0$, row sum

Fig. 13 Histograms for average row sum of the successful targeted attack of each character with different perturbation value ($\delta>0$)

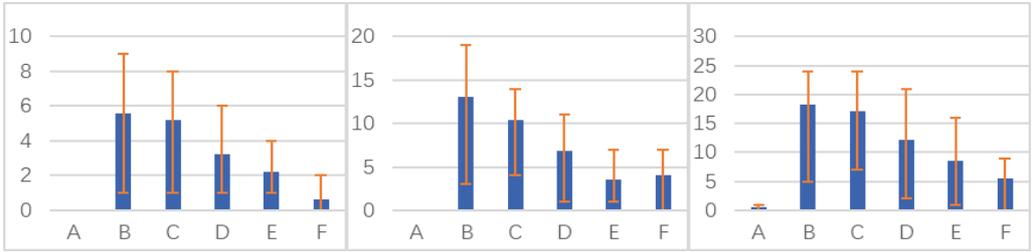

(a) 3×3, $\delta<0$, row sum    (b) 5×5, $\delta<0$, row sum    (c) 7×7, $\delta<0$, row sum

Fig. 14 Histogram for average row sum of the successful targeted attack of each character with different perturbation value ($\delta<0$)

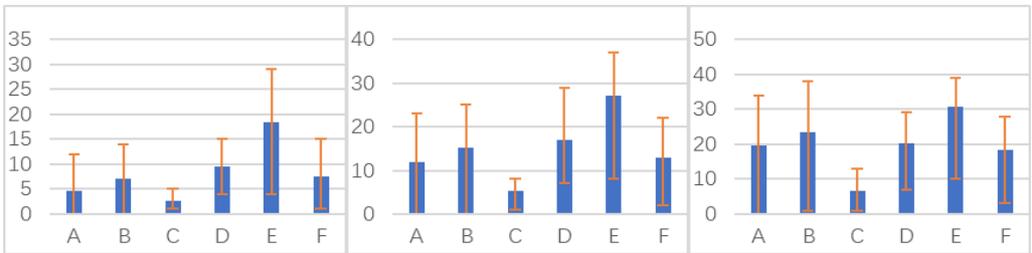

(a) 3×3, $\delta>0$, column sum    (b) 5×5, $\delta>0$, column sum    (c) 7×7, $\delta>0$, column sum

Fig. 15 Histogram for average column sum of the successful targeted attack of each character with different perturbation value ($\delta>0$)

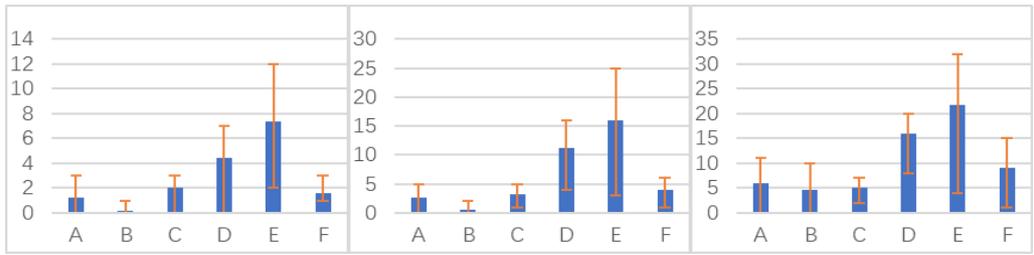

(a) 3×3, $\delta<0$, column sum　　(b) 5×5, $\delta<0$, column sum　　(c) 7×7, $\delta<0$, column sum

Fig. 16 Histogram for average column sum of the successful targeted attack of each character with different perturbation value ($\delta<0$)